# A Three-Phase Artificial Orcas Algorithm for Continuous and Discrete Problems

Habiba Drias, University of Science and Technology Houari Boumediene, Algeria*

Lydia Sonia Bendimerad, University of Science and Technology Houari Boumediene, Algeria

Yassine Drias, University of Algiers, Algeria

https://orcid.org/0000-0002-8896-6170

## ABSTRACT

In this paper, a new swarm intelligence algorithm based on orca behaviors is proposed for problem solving. The algorithm called artificial orca algorithm (AOA) consists of simulating the orca lifestyle and in particular the social organization, the echolocation mechanism, and some hunting techniques. The originality of the proposal is that for the first time a meta-heuristic simulates simultaneously several behaviors of just one animal species. AOA was adapted to discrete problems and applied on the maze game with four level of complexity. A bunch of substantial experiments were undertaken to set the algorithm parameters for this issue. The algorithm performance was assessed by considering the success rate, the run time, and the solution path size. Finally, for comparison purposes, the authors conducted a set of experiments on state-of-the-art evolutionary algorithms, namely ACO, BA, BSO, EHO, PSO, and WOA. The overall obtained results clearly show the superiority of AOA over the other tested algorithms.

## INTRODUCTION AND MOTIVATION

Based on the *No Free Lunch Theorem* (Adam & Alexandropoulos, 2019), a swarm intelligence algorithm integrating several important behaviors from animal intelligence was designed. Such rich comportments are found in Orcas, which are marine mammals belonging to the delphinidae family. Their lifestyle is simulated by an algorithm integrating several intelligent behaviors such as social organization, echolocation phenomenon and hunting techniques used for the detection of the prey.

The first premises of the proposal have been already previewed under an algorithm called OA (Orcas Algorithm) (Bendimerad, Drias, 2021). This paper has profoundly reshaped the initial idea both on the theoretical development level and on the experimental side.

The proposed algorithm, called AOA (Artificial Orca Algorithm) is developed by simulating the mentioned orcas' skills in a deeply and detailed way. First, the orcas make a simple search for prey based on swarm intelligence concept during their journey. Second, if they do not find easy prey on their way, they employ more expensive techniques, first echolocation to detect the prey and then hunting if it turns out that the echolocation is not sufficient.

A discrete version is designed and applied to the problem of finding the optimal path in labyrinths. Further, a comparative study is performed to show AOA effectiveness and efficiency relatively to state-of-the-art algorithms, namely, ACO, BA, BSO, EHO, PSO and WOA.

This article is organized around six main sections. The next section presents a summarized state-of-the-art on swarm algorithms. the third is a brief presentation of the social organization of the orcas. The fourth describes step-by-step the design of the proposed algorithm through the modeling of the orca lifestyle including the swarm collective intelligence, the echolocation, the hunting techniques and the exploration phase. The fifth section shows how to adapt AOA to combinatorial optimization problems such as the maze game. Experiments on public datasets representing mazes and an analysis of the obtained results are exhibited. A comparative study with state-of-the-art algorithms is also performed. The last section concludes the contributions of this study and points out some future works.

## SOCIAL ORGANIZATION OF THE ORCAS

The orcas' society, generally run by elderly women (matriarchs), consists often of large groups of family members that may span several generations (Killer-Whale, 2022). The resident orcas (OrcaLab, 2022) organize themselves into a matrilineal structure. For this type of orcas, the mother and her adult sons stay together all their lives. Nevertheless, girls can take their independence as soon as they have a descent, but nothing prevents them from seeing each other from time to time. The resident orcas' society is structured hierarchically as follows:

*A matrilineal unit.* It consists of the matriarch and her descendants (on average four generations). The stability of these groups is seen in their rare inseparability.
*A pod.* It is composed of two to four matrilineal units, consisting of about 20 individuals, constituting an extended family of close mothers. These groups may separate for a period before the reunion. A pod can be defined as the orcas are usually seen traveling together.
*A clan.* It represents the third social level, it includes a set of pods, which share the same dialect and a distant ancestor.
*A Community.* It is the last and highest social level, composed of clans that socialize and meet regularly, but do not share the same dialect or common ancestors.

Figure 1 depicts this specific organization.

**Figure 1. The social orcas' organization**

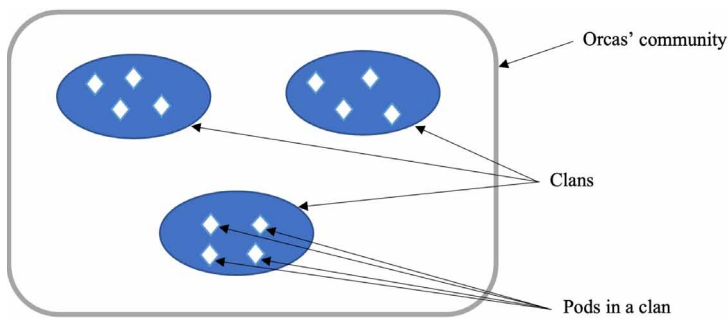

## BACKGROUND

Problem-solving algorithms and in particular evolutionary algorithms are inspired partly from nature and biology. The Earth contains a variety of animals that continually surprise scientists by their intelligence. In the domain of swarm intelligence, researchers aim at establishing algorithms that simulate their intelligence and way of life. In the beginning, the research field focused on simulating social insects' behaviors to translate the group intelligence. Algorithms such as ACO (Ant Colony Optimization) (Dorigo & Stutzle, 2006) were published, followed by bacterial growth algorithm (Passino, 2002). Lately, birds and fishes' behaviors were investigated for developing another type of intelligent algorithms such as PSO (Particle Swarm Optimization) (Clerc, 2013; Oliveira & Pinheiro, 2016), BA (Bat Algorithm) (Yang & He, 2013) and FFA (Fire Fly Algorithm) (Yang & He, 2013).

Recently, animal herding has attracted researchers interest. Algorithms such as EHO (Elephant Herding Optimization) (Wang & Dos Santos Coelho, 2016), ESWSA (Elephant Swarm Water Search Algorithm) (Mandal, 2018) and WOA (Whale Optimization Algorithm) (Mirjalili & Lewis, 2016) were designed.

Several applications of these algorithms were performed in various domains. For instance, ACO was used to solve scheduling of tasks while preventing the premature convergence. BSO has addressed the graph coloring issue and the web information foraging issue. PSO was widely utilized as it is simple to implement. WOA has been shown to perform efficiently on neural network connection weights (Aljarah et al., 2018), image segmentation (Aziz et al., 2017) and feature selection (Mafarja & Mirjalili, 2017). EHO was applied to optimize controller parameters (Gupta & Singh, 2016), appliance scheduling (Mohsin & Javaid, 2018), support vector machine parameters (Tuba, 2017), image processing (Tuba & Alihodzic, 2017), community detection (Belkhiri et al., 2022) and Target Detection in Complex and Unknown Environments (Houacine & Drias, 2021). BA, which can be seen as a generalization of PSO was also utilized widely in disciplines such as computer vision (Akhtar & Ahmad, 2019), association rule mining (Heraguemi et al. 2018), information retrieval (Khennak & Drias 2017), data classification (Mishra & Bande, 2008) and data clustering (Sood & Bansal, 2013).

## THE ARTIFICIAL ORCA ALGORITHM (AOA)

To reach the prey, which is of various marine species, resident orcas use very diverse techniques. They catch the easy prey just by moving using their collective intelligence and the tough one by using echolocation system before hunting. These major behaviors are simulated in a 3-phase algorithm called AOA (Artificial Orca Algorithm). The social motion, the echolocation mechanism and the hunting techniques are simulated for the exploitation phase of the search process. The exploration phase simulates the comportment of females who take their independence as soon as they have a descent, while keeping the family ties. These phases are outlined in the next subsections.

### Modeling Collective Motion

Based on swarm principles and in order to preserve their social organization, the orcas move in such a way to avoid collisions with nearby flock mates, to align velocity with that of nearby flock mates and to stay close to flock center. Therefore, the movement of an orca depends on its current speed, on the best performance of the neighbors of the pod to which it belongs and on the best performance of the clan to which it belongs. This phenomenon is simulated in an algorithm for searching optimal solutions in a huge search space of potential solutions. Each single solution simulates an orca with a position at a given time aiming at evolving towards an optimal solution. For that purpose, it is characterized by a velocity that directs its motion and is evaluated by a fitness function that measures its performance consisting in appreciating its proximity with an optimal solution. The process is initialized with a group of artificial orcas (solutions or individuals) drawn at random with the objective of seeking for the best solution (which corresponds to the position of the prey in the real world). It renews the search for many iterations until finding an optimal solution or an approximate solution with a high quality. At each iteration, each orca updates its speed and its position relatively to its pod mates and clan mates in order to follow its congeners in the swarm. Let $x^*_{p_i}$ be the best position reached so far in pod *pi* where orca *i* resides and $x^*_{c_i}$, the best solution obtained so far by any orca in clan $c_i$ where orca *i* exists. If $v_i^{t-1}$ is the velocity of orca *i* at time $t-1$, then its update is expressed by Equation 1.

$$v_i^t = w_i v_i^{t-1} + w_{p_i} * d\left(x^*_{p_i}, x_i^{t-1}\right) + w_{c_i} * d\left(x^*_{c_i}, x_i^{t-1}\right) \tag{1}$$

where $d(x, y)$ is the distance that separates position *x* from position *y*. $w_i$ is an empirical parameter denoting the move inertia weight of orca *i*. $w_{p_i}$ is the orca social skill in pod $p_i$ it belongs to. It simulates the contribution of the orca for the pod and is calculated as shown in Equation 2. $w_{c_i}$ is the orca social skill in clan $c_i$ it belongs to and is calculated by Equation 3 as the contribution of the individual for the clan. $f(x)$ is the fitness function measuring the performance of the solution *x*.

$$w_{p_i} = \frac{f\left(x_i^{t-1}\right)}{\sum_{j \in p_i} f\left(x_j^{t-1}\right)} \tag{2}$$

$$w_{c_i} = \frac{f\left(x_i^{t-1}\right)}{\sum_{j \in c_i} f\left(x_j^{t-1}\right)} \tag{3}$$

The position $x_i^t$ of orca *i* at time *t* is then expressed by Equation 4.

$$x_i^t = v_i^t + x_i^{t-1} \tag{4}$$

In order to take into account, the water wave motion, the individual will undergo a wave move (Craik, 2004) expressed by Equation 5.

$$x_i^t = \gamma * (\sin \frac{2\pi}{L} x_i^t - \frac{2\pi}{L} t) \tag{5}$$

where:

- $\gamma$ is an empirical parameter that denotes the wave amplitude.
- L and T represent respectively the wave length and the wave period. It is assumed that a wave lasts during one iteration of the search process. Hence $T = 1$ and $L = s * T = s$ where *s* is the speed of sound, which is equal to 1481 meters per second.
- *t* is the time the individual spends to cross the wave during one iteration, which is expressed as $t = \frac{L}{v_i^t}$.

If this phase fails to meet an optimal solution, then the echolocation phase described in the next subsection is launched.

## Modeling Echolocation

Except when tracking other cetaceans, orcas frequently use echolocation which consists of an emission of a very strong sound impulse and the return of an echo allowing the perception of the surrounding objects (Sea-World 2022). The sound crosses the water and brings vibrations back with valuable information about the prey, notably its size and its proximity. As orcas use a high speed in the echolocation mechanism, the velocity of the orca is modeled according to this characteristic. Each individual *i* of the population computed and returned by the first phase at time *t* is assigned a frequency $f_i^t$ and a loudness $A_i^t$. The frequency varying between two empirical parameters $f_{min}$ and $f_{max}$ is expressed using Equation 6, where $\beta$ is a random value in the range [0, 1]. The loudness emitted by the orca is very strong at the beginning of the echolocation, then it decreases as it gets close to the prey. It varies from a positive large value $A_0$ to a minimum value $A_{min}$ formulated by Equation 7, where $\delta$ is an empirical parameter from the range [0, 1] making $A_i^t$ decreasing at each iteration. The velocity follows the same rhythm as the loudness and is expressed by Equation 8. Equations 6, 7, 8 and 9 are then calculated for many iterations to improve the matriarch position. If the echolocation stage is not sufficient to encounter an optimal solution then, the hunting process is launched.

$$f_i^t = f_{min} + \left(f_{max} - f_{min}\right)\beta \tag{6}$$

$$A_i^t = \delta A_i^{t-1} \tag{7}$$

$$v_i^t = \frac{1}{f_i^t} A_i^t \tag{8}$$

$$x_i^t = v_i^t + x_i^{t-1} \tag{9}$$

## Modeling Hunting Techniques

The range of hunting techniques developed by the orcas is vast, and depends on both the prey and the environment (Dutfield, 2022). Scientists have identified at least six kinds of them, namely, the *Wave Wash* for attacking seals, the *Karate Chop* for sharks, the *Carousel* for small fish, the *Pod Pin* for narwhals, the *Blowhole Block* for larger cetaceans such as humpback whales and the *D-Day* for sea lions and elephant seals. The population generated by the echolocation phase initiates the hunting process. For the simulation of this phenomenon, the focus is on the so-called *Carousel* hunting technique, which can be extended to other techniques mentioned above. The strategy consists in swimming around the prey on a circle while shrinking it at each iteration in order to reach the target. The other way is the use of a spiral path to catch the prey.

### The Narrowing Encircling Mechanism

In this case, the orcas move on a circle while decreasing its radius at each iteration, shrinking this way the distance separating them from the prey. Equation 10 first positions the orcas on a circle whose center is the position of the current matriarch and the radius a random number $r$. $d\left(x_i^{t-1}, x^*\right)$ is the distance that separates the individual $i$ from the population matriarch at time $t-1$. If the position is outside the circle, then the first formula of the bracket is applied otherwise the second one is considered. Then by using Equation 11, which reduces at each iteration the radius by a constant $a$, the orca moves in the direction of the prey, $a$ being a random fraction of $r$.

$$x_i^t = \begin{cases} x_i^{t-1} + d\left(x_i^{t-1}, x^*\right) - r & \text{if } d\left(x_i^{t-1}, x^*\right) > r \\ x_i^{t-1} + r - d\left(x_i^{t-1}, x^*\right) & \text{otherwise} \end{cases} \qquad (10)$$

$$x_i^t = x_i^{t-1} + r - a \qquad (11)$$

### The Spiral Model

Equation 12 simulates a simple Archimedes spiral movement where $l$ is a random number. But prior to proceeding with this movement, the orcas should position themselves at the same distance from the matriarch using Equation 10.

$$x_i^t = x_i^{t-1} - 2\pi l \qquad (12)$$

Then it is assumed there is a probability $\alpha$ of choosing between the narrowing encircling mechanism and the spiral model to update the position of the orcas, which gives rise to Equation 13.

$$x_i^t = \begin{cases} x_i^{t-1} + r - a & \text{if } p < \alpha \\ x_i^{t-1} - 2\pi l & \text{otherwise} \end{cases} \qquad (13)$$

### Exploration Strategy

Inspired by the behavior of resident female orcas leaving the family group, the exploration strategy draws at random a matriarch from the clans, to let it move as far as possible according to its abilities, aiming at creating a totally new population different from the previous ones. The exploration of a totally new region of the search space for several times, guaranties at the end a quasi-comprehensive exploration of the search space and hence prevents from premature convergence. Let $x^*$ be the selected matriarch, then Equation 14 formulates its new position $x_{new}^*$, $MaxDistance\left(x^*\right)$ is the distance the matriarch can run through, according to its capacity.

$$x_{new}^* = x^* + MaxDistance\left(x^*\right) \qquad (14)$$

## AOA IMPLEMENTATION

In this section, the Artificial Orca Algorithm (AOA) based on the lifestyle of resident orcas characterized by the behaviors mentioned above, is proposed. The search space of a problem to be solved corresponds to the set of all potential solutions that correspond to positions that artificial orcas can reach. An optimal solution in the search space represents the position of the prey in the real world. A population of orcas consists of clans that stand out from each other, where each clan contains a number of pods, which at their turn contain individuals. There is a decreasing order between the distances separating clans, pods and individuals respectively. The distance separating clans is greater than the distance that separates pods of the same clan, which is greater than the distance that separates individuals of the same pod. AOA and its three phases are outlined respectively in Algorithms 1, 3, 4 and 5.

**Algorithm 1.** AOA (Artificial Orca Algorithm)
**Require:** Empirical parameters; A taboo list set to empty;
**output:** optimal or approximate solution;
begin
draw at random an individual *i*;
create-population of *i*;
**for** MaxIter **do**
Calculate the fitness value of each individual;
Determine the matriarch of each pod and each clan;
Collective-motion-search;
**if** (optimal solution **not** found) **then** Echolocation; **end if**;
**if** (optimal solution **not** found) **then** Hunting; **end if**;
**if** (optimal solution **not** found) **then**
draw at random a matriarch among those of the clans;
apply Equation (14);
**while** the new position in taboo list **do** apply Equation (14);
insert the new position in the taboo list;
generate a population for the new position;
end while;
end if;
end for;
end.

In Algorithm 2, a population of individuals is created according to the social structure of the orcas. The time complexity of the algorithm is $O(n_{i,p} * n_{p,c} * n_c)$, where $n_{i,p} * n_{p,c} * n_c$ is equal to *n*, the total number of individuals in the population called also the population size. Then, the complexity can be simply expressed as $O(n)$.

In Algorithm 3, each individual undergoes a position change for *MaxMotions* times. The complexity of this operation is then $O(n*MaxMotions)$. The update of the matriarch corresponds to the search of the individual with the highest fitness value. Its complexity is then $O(n)$ for determining the matriarchs of all the pods and $O(n_{p,c} * n_c)$ for determining the matriarchs of the clans. Therefore, the complexity of this algorithm is $O(n*MaxMotions)$.

Similarly, the respective complexities of Algorithms 4 and 5 are equal to $O(n*MaxEchoMotions)$ and $O(n*MaxHuntMotions)$. The complexity of the exploration phase is $O(n)$ as it draws at random a matriarch of a clan and creates a population starting with this individual.

The worst-case complexity of one iteration of Algorithm 1 is the sum of the complexities of Algorithms 2, 3, 4, 5 and the exploration phase, as these steps are called sequentially.

It is then equal to $O(n + n*MaxMotions + n*MaxEchoMotions + n*MaxHuntMotions + n) = O(n*(MaxMotions + MaxEchoMotions + MaxHuntMotions))$.

The complexity of AOA is then equal to $O(n*(MaxMotions + MaxEchoMotions + MaxHuntMotions)*MaxIter)$. The algorithm has a polynomial complexity that depends on the population size, the number of global iterations and the number of iterations of each phase.

## DISCRETE AOA AND APPLICATION FOR THE MAZE PROBLEM

In this section, the discrete variant of AOA is particularly studied through its application to the maze game. A labyrinth is represented by a two-dimensional rectangular grid of size $n*m$, with an entrance and an exit. It consists of cells that can contain obstacles. The link between the obstacles in the labyrinth represents the notion of connectivity (Alaguna & Gomez, 2018). A solution for the maze problem represents a set of moves established by a moving agent from the entrance to the exit. The possible moves are the four directions: *left*, *right*, *up* and *down*. A solution is then represented by a vector of moves and characterized by a length as shown in Table 1.

**Table 1. An example of a solution of length equal to 8**

| left | right | up | right | up | up | up | right |
|---|---|---|---|---|---|---|---|

If a maximum length *l* is assigned to a solution, then the size of the search space would be equal to $4^l$, which is huge when the value of *l* is large. If *l* is at least equal to 10 for instance, then the search space will include more than $10^6$ potential solutions. The use of meta-heuristics to address the maze problem is therefore beneficial.

The changes made to AOA to be applicable to discrete problems only affect the computation of the different equations. Adding a number *k* to a solution consists in adding *k* random moves to the corresponding path while respecting the admissibility of the solution. If no such construct is possible, that is, if the agent cannot move forward, then a backtracking is allowed. Subtracting a number *k* from a solution consists in removing the *k* last moves from the solution. The multiplication of a solution by a number *k* is equivalent to appending to the solution, a sequence of $(k-1)$ solution paths.

The fitness function of a solution is the Manhattan distance between the last cell reached by the process and the exit cell.

### Generation of the Initial Population

Clans of one population share some features (moves in our case) between themselves while remaining distant from each other. One idea to create clans is to consider an initial path and to cut it in parts to initialize the clans. As cut-points *cp*, the positions computed by Equation (15) are proposed, where the division yields an integer number. *sol-size* is the solution path size and *#clans* is the number of clans.

$$cp = \begin{cases} \dfrac{sol-size}{\#clans-1} + 1 & if\ sol-size \geq clans\ \#\ \#\#\#\# \\ \dfrac{sol-size}{\#clans-1} + 1 & otherwise \end{cases} \quad (15)$$

For instance, let consider the previous solution and a number of clans equal to 3 then the first clan will be represented as shown in Table 2.

The process is then launched to create the second clan shown in Table 3,

and repeated to yield the third clan as shown in Table 4.

Table 2. Representation of the first clan

| left | right | up | right | up | up | up | right |
|------|-------|-----|-------|-----|-----|-----|-------|

Table 3. Representation of the second clan

| left | right | up | right | up |
|------|-------|-----|-------|-----|

Table 4. Representation of the third clan

| left | right | up |
|------|-------|-----|

Then from each clan, pods are created by adding a fixed number of moves to build an initial solution for its individuals. Subsequently individuals from the same pod will share the same starting location to start their search. Besides, for reasons of simplification, the distance $d_c$ between the clans is set to the size of the initial solution, the distance $d_p$ between the pods of the same clan is set to 2/3 of the size of the initial solution and the distance $d_i$ between the individuals of the same pod is set to 1/3 of the size of the initial solution.

## EXPERIMENTS

The dataset used to evaluate AOA, represents a set of labyrinths, available on gitHub (Maze, 2020). In order to evaluate the ability of our algorithm on several types of labyrinths, similarly connected maze problems (SCMP) containing four types of labyrinths are used: SCMP1 with a connectivity equal to 0%, SCMP2 with a connectivity equal to 30%, SCMP3 with a connectivity equal to 60%, and SCMP4 with a connectivity equal to 100%. Each type contains 10 labyrinths and 12 starting positions, which gives 120 instances for each type and thus a total number of instances equal to 480 instances.

## Parameters' setting

In the first bunch of experiments dealing with parameters tuning, the population size (#*Clans*, #*Pods*, #*Individuals*) was varied as described in Table 5, in order to have a population description fairly close to the natural orcas' population.

Table 5. Configurations used to tune the population size

|      | #clans | #Pods | #Individuals | Population size |
|------|--------|-------|--------------|-----------------|
| Pop1 | 2      | 2     | 5            | 20              |
| Pop2 | 2      | 2     | 10           | 40              |
| Pop3 | 2      | 4     | 5            | 40              |
| Pop4 | 2      | 4     | 10           | 80              |
| Pop5 | 4      | 2     | 5            | 40              |
| Pop6 | 4      | 2     | 10           | 80              |
| Pop7 | 4      | 4     | 5            | 80              |
| Pop8 | 4      | 4     | 10           | 160             |

From Figure 2, the best success rate achieved by the 4 types of mazes is that of *Pop5*, where *#clan=4*, *#pod=2* and *#individuals=5*, with a population size = 40. Also, this configuration consumes the minimum amount of execution time.

Figure 2. Statistics results of AOA when varying the population structure

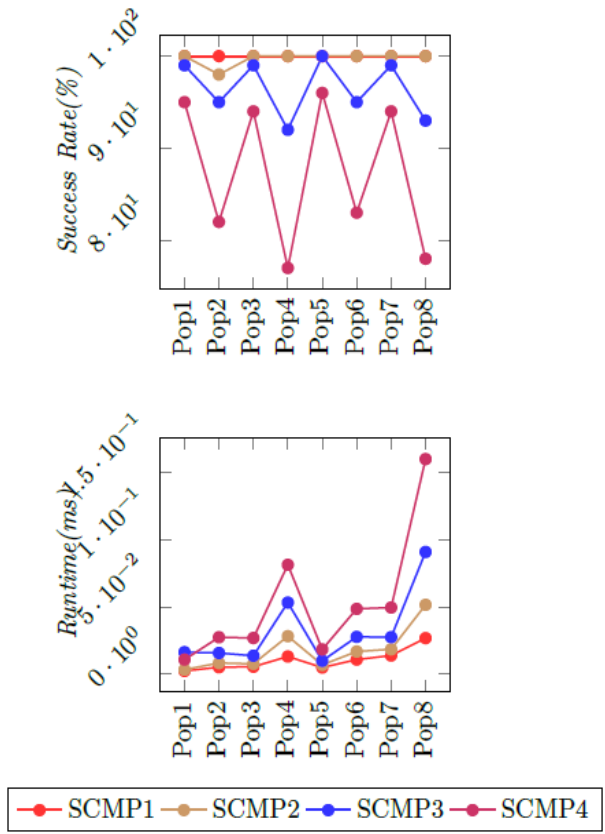

After fixing the population size, the number of iterations, *MaxIter*, *MaxMotions*, *MaxEchoMotions* and *MaxHuntMotions* as reported in Table 6 was varied. In order these parameters can fit the table, they are represented respectively by *G-Iter*, *M-Iter*, *E-Iter* and *H-Iter*.

Figure 3 depicts the results of AOA on the different configurations. The iteration combination corresponding to *Iter5* reaches 100% of the success rate. The average run time for this configuration is the lowest for SCMP1, SCMP2 and SCMP3 and among the lowest for SCMP4. Therefore, *MaxIter*, *MaxMotions*, *MaxEchoMotions* and *MaxHuntMotions* were set respectively to 5, 50, 30, 30 with a total number of iterations equal to 550.

Figures 4, 5, 6, 7 and 8 expose respectively the results when varying the frequencies *fmin and fmax* in {(0,1),(1,2),(0,2)}, $\gamma$, $\delta$, and $w_0$ in {0.25, 0.5, 0.75}. The best values for *fmin and fmax* are (0, 2), where the success rate reaches 100% and the run time has almost the same magnitude as that for the other configurations.

The parameter $\gamma$ was set to 0.5 even if there is not a big difference in the outcomes produced by $\gamma = 0.5$ *and* 0.75. However, the value 0.5 yields a better solution size. For $\delta$, the values 0.25 *and*

Table 6. Configurations used to tune the number of iterations

|        | G-Iter | M-Iter | E-Iter | H-Iter | Total |
|--------|--------|--------|--------|--------|-------|
| Iter1  | 5      | 30     | 30     | 30     | 450   |
| Iter2  | 5      | 30     | 30     | 50     | 550   |
| Iter3  | 5      | 30     | 50     | 30     | 550   |
| Iter4  | 5      | 30     | 50     | 50     | 650   |
| Iter5  | 5      | 50     | 30     | 30     | 550   |
| Iter6  | 5      | 50     | 30     | 50     | 650   |
| Iter7  | 5      | 50     | 50     | 30     | 650   |
| Iter8  | 5      | 50     | 50     | 50     | 750   |
| Iter9  | 10     | 30     | 30     | 30     | 900   |
| Iter10 | 10     | 30     | 30     | 50     | 1100  |
| Iter11 | 10     | 30     | 50     | 30     | 1100  |
| Iter12 | 10     | 30     | 50     | 50     | 1300  |
| Iter13 | 10     | 50     | 30     | 30     | 1100  |
| Iter14 | 10     | 50     | 30     | 50     | 1300  |
| Iter15 | 10     | 50     | 50     | 30     | 1300  |
| Iter16 | 10     | 50     | 50     | 50     | 1500  |

Figure 3. Statistics results of AOA when varying the iterations structure

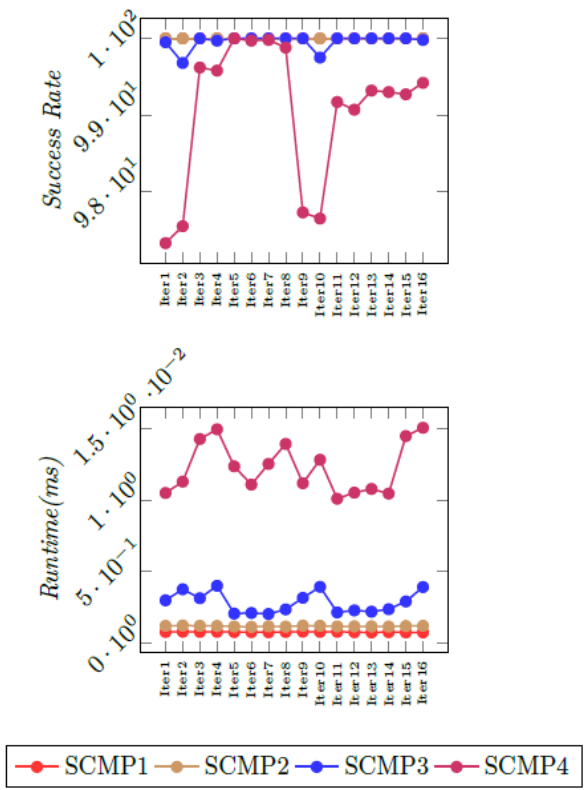

**Figure 4. Statistics results of AOA when varying frequencies**

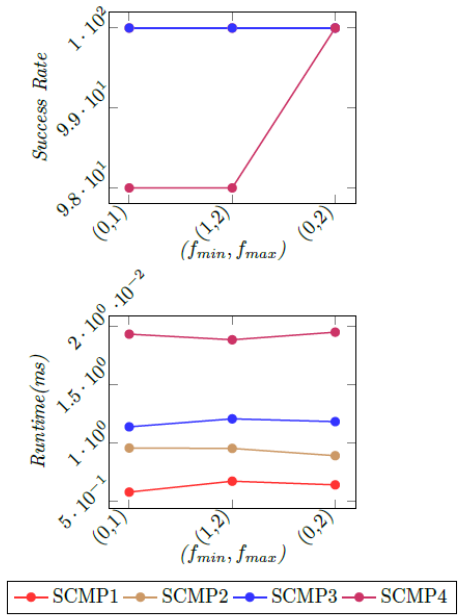

**Figure 5. Tuning $\gamma$ parameter**

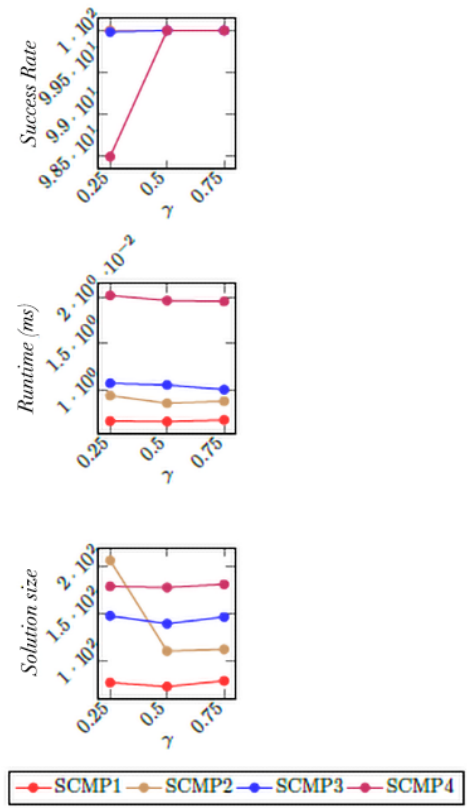

**Figure 6. Tuning $\delta$ parameter**

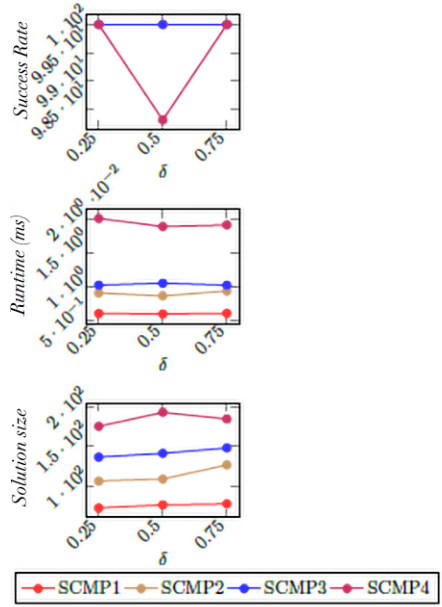

**Figure 7. $\alpha$ parameter tuning**

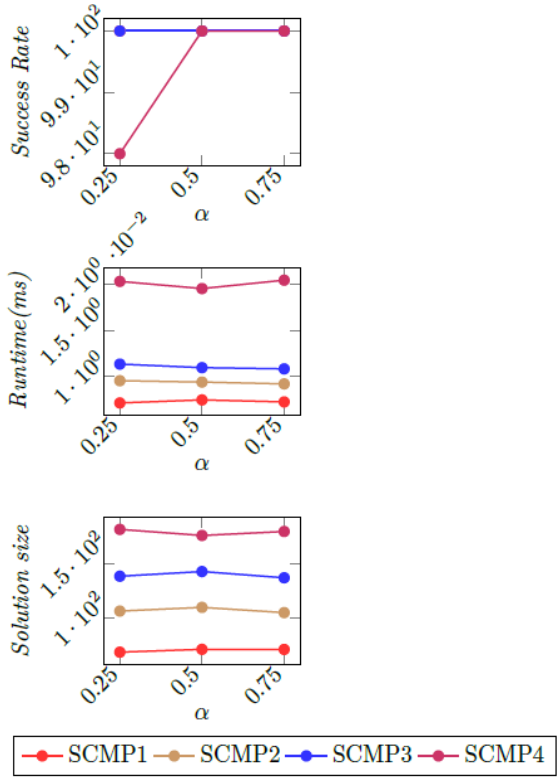

**Figure 8. w0 parameter tuning**

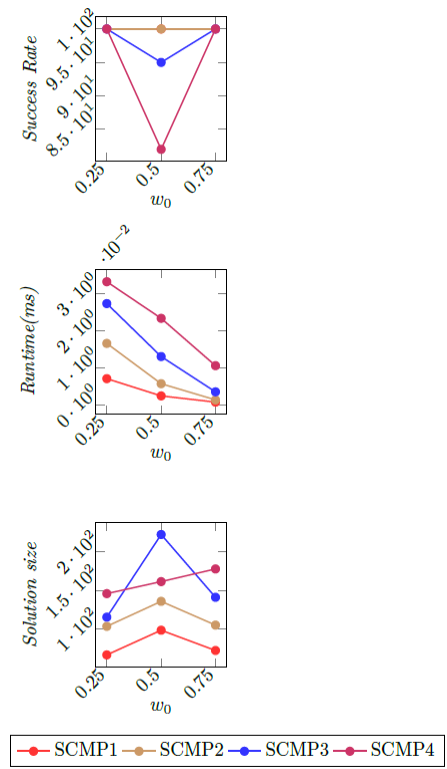

0.75 reach 100% of success rate. The tie is broken in favor of $\delta = 0.25$ that yields the best solution size. For the parameter $\alpha$, it is clear that the best value is 0.75 according to the results. Finally, the value attributed to $w0$ is 0.25 according to the average success rate and the average solution size, even if the run time is longer compared to the other cases.

## AOA Performance Evaluation

To determine the performance of *AOA*, three statistics, namely, the average success rate, the average run time and the average solution size are computed over 50 runs. Note that the optimal solution corresponds to a success rate of 100% and to the smallest solution size.

## Results for 15x15 Labyrinths

For all tested 15x15 labyrinths, AOA reaches an optimal solution with 100% success rate. Figures 9 and 10 present respectively the average solution size and the average execution time of each type of labyrinth. let notice that AOA is able to solve all kinds of 15X15 in a reasonable time and that SCMP4 requires more moves to reach an optimal solution relatively to the other types of labyrinths.

## Results for 30x30 Labyrinths

Figures 11 and 12 depict respectively the average solution size and the average run time for each type of 30x30 labyrinths. Let observe that it is easy for AOA to reach the optimal solution for SCMP1, SCMP2 and SCMP3 mazes but it needs relatively more time to reach the optimal solution for SCMP4 instances. The best average solution size is higher for SCMP4 relatively to the other types of mazes.

**Figure 9. Average solution size obtained by AOA for each complexity degree of 15x15 maze**

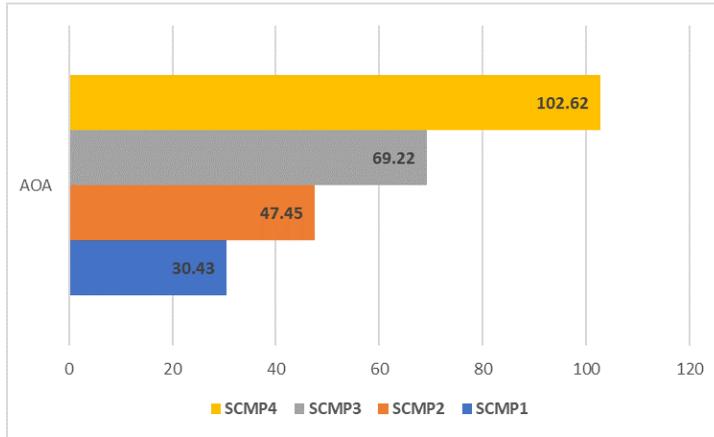

**Figure 10. Average run time obtained by AOA for each complexity degree of 15x15 maze**

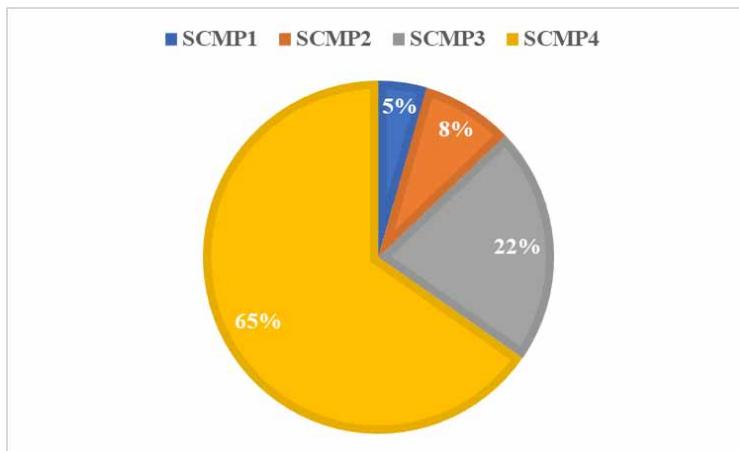

**Figure 11. Average solution size for 30x30 mazes**

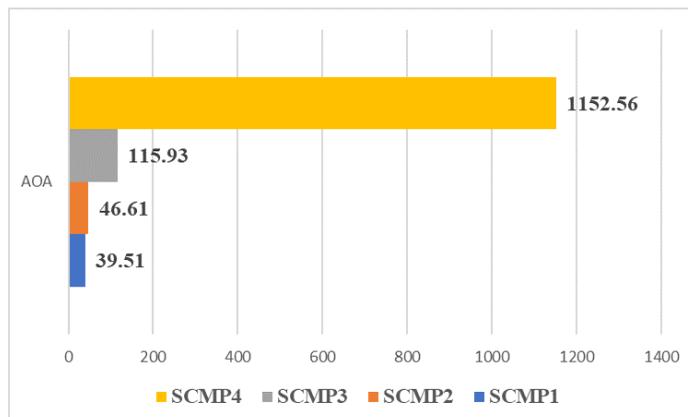

**Figure 12. Average runtime for 30x30 mazes**

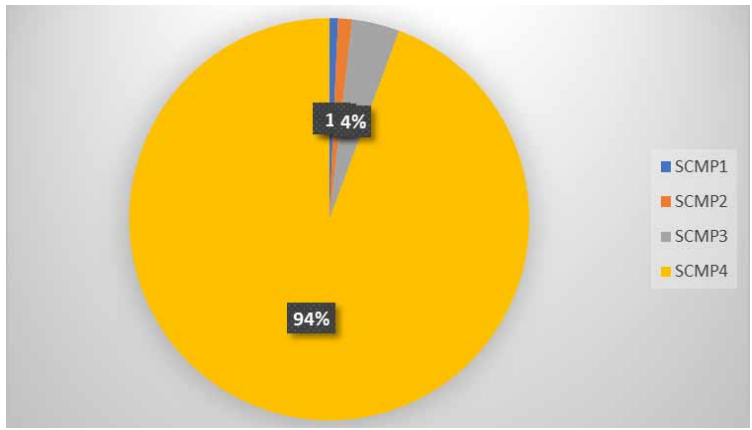

From these set of experiments, the conclusion is that tackling the maze problem depends on the maze size and on its structure complexity. Higher are its size and its complexity, longer are the solution length and the running time.

## Comparison with State-of-the-art Algorithms

Classic state-of-the-arts algorithms namely ACO, BSO and PSO and recent ones such as BA, WOA and EHO are tested and compared to AOA as they have proven their ability to solve hard problems. These algorithms were adapted to maze problem in the same way as AOA. Table 7 exhibits the average success rate (SR), the average solution size (SS) and the average run time (RT) calculated by these methods for the four types of labyrinth. Note that the solution size is reported only when the success rate reaches 100%. Figures 13 and 14 illustrate these numerical results for a better analysis of the outcomes. Let observe that the only algorithms that are able to solve the problem with 100% success rate for all types of labyrinth are AOA, BA, BSO and PSO. BSO presents the disadvantage of being slow relatively to the other three methods. Among the remaining best algorithms, that is, AOA, BA and PSO, AOA has the best performance as it yields the smallest average solution size.

**Table 7. Comparing AOA with 6 other swarm intelligence algorithms on the maze problem**

|  |  | ACO | AOA | BA | BSO | EHO | PSO | WOA |
|---|---|---|---|---|---|---|---|---|
| SCMP1 | SR | 100% | 100% | 100% | 100% | 100% | 100% | 83% |
|  | SS | 13.50 | 30.43 | 30.63 | 43.34 | 19.57 | 34.59 | - |
|  | RT | 0.0013 | 0.0010 | 0.0023 | 0.4824 | 0.0045 | 0.0077 | 0.0358 |
| SCMP2 | SR | 100% | 100% | 100% | 0.482371 | 76% | 100% | 60% |
|  | SS | 44.24 | 47.45 | 45.79 | 47.12 | - | 49.57 | - |
|  | RT | 0.0073 | 0.0019 | 0.0033 | 1.0615 | 0.0113 | 0.0063 | 0.0726 |
| SCMP3 | SR | 86% | 100% | 100% | 100% | 65% | 100% | 50% |
|  | SS | - | 69.22 | 70.31 | 44.85 | - | 68.73 | - |
|  | RT | 0.0255 | 0.0049 | 0.0045 | 0.9897 | 0.0165 | 0.0067 | 0.0664 |
| SCMP4 | SR | 75% | 100% | 100% | 100% | 67% | 100% | 41% |
|  | SS | - | 102.62 | 243.23 | 49.46 | - | 110.37 | - |
|  | RT | 0.0425 | 0.0147 | 0.0235 | 1.0174 | 0.0178 | 0.0056 | 0.0682 |

**Figure 13. Success rate and runtime of AOA and six other swarm algorithms on the maze problem**

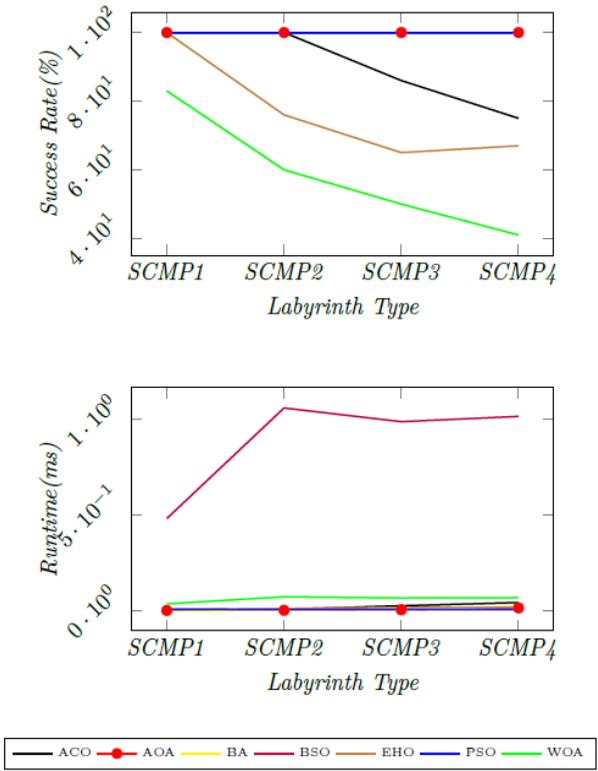

**Figure 14. Average solution size of AOA and six other swarm algorithms on the maze problem**

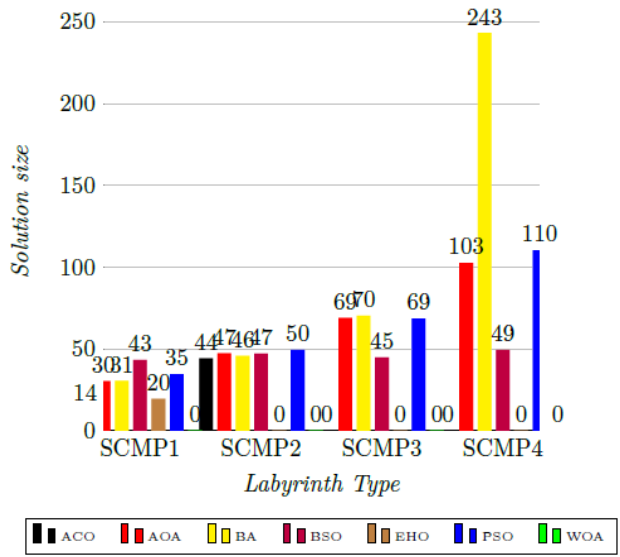

## CONCLUSION

Impressed by the social and cognitive skills of orcas, making them very intelligent beings, this study is dedicated to translate their intelligence into a sophisticated solving problem tool. Their group movement based on collective intelligence, their mastery of the echolocation phenomenon and of various hunting techniques classify them at the top of the predators' hierarchy. These characteristics are modeled for the design of an algorithm called AOA (Artificial Orcas Algorithm). First, the collective intelligence is simulated in an original way, with respect to the orcas' social organization. Second, the echolocation technique reflects faithfully the natural phenomenon. Third, the modeling of the *Carousel* strategy of hunting, uses as the previous phases, mathematical developments for describing the metaphor. In addition, the computational complexity of each phase is provided prior to yielding that of the global algorithm.

From the experimental point of view, AOA was adapted to discrete problems and was tested on public datasets related to the maze game with four types of complexity, namely, SCMP1, SCMP2, SCMP3 and SCMP4. The complexity of the search space for the maze game was argued theoretically. Extensive experiments were performed to tune the various parameters of the algorithm for two different dimensions of maze, which are 15x15 and 30x30. The algorithm performance was highlighted by exhibiting the average success rate, the average solution size and the average execution time. Afterwards, for comparatives, six state-of-the-art algorithms WOA (Whale Optimization Algorithm), EHO (Elephant Herding Algorithm) and BA (Bat Algorithm) and Classic approaches ACO (Ant Colony Optimization), BSO (Bee Swarm Optimization) and PSO (Particle Swarm Optimization), were implemented for the maze game. The outcomes show that AOA yields the best success rate, the best run time and the best solution size for the labyrinth problem.

These findings reflect somehow the superiority of the behavior of orcas as it is richer and more dynamic than that of the other species compared with. Indeed, ACO, BSO and PSO simulate only the collective intelligence. BA incorporates no particular hunting techniques; its strength resides only in the echolocation experience. Let also note the absence of hunting techniques modeling in EHO. Finally, in addition to the echolocation phenomenon, AOA has a very sophisticated population organization and hunting techniques that do not exist in WOA.

The results obtained are very encouraging to adapt the proposed discrete algorithm to other issues. Also, in rder to improve the efficiency of the algorithm to reach scalability, an implementation on GPU is planned, as the community structure of the orcas lends itself to parallel computation. Another promising but tricky research axis would be to explore the hunting strategies in depth and model them in order to enrich the last phase of the algorithm.

## FUNDING AGENCY

The Open Access Processing fee for this article has been waived by the publisher.

## ACKNOWLEDGMENT

This research was supported by the Directorate General for Scientific Research and Technological Development (DGRSDT) [grant number C0662300].

# REFERENCES

Adam, S. P., Alexandropoulos, S. N., Pardalos, P. M., & Vrahatis, M. N. (2019). No Free Lunch Theorem: A Review. In I. Demetriou & P. Pardalos (Eds.), *Approximation and Optimization, Springer Optimization and Its Applications* (Vol. 145, pp. 57–82). Springer.

Akhtar, S., Ahmad, A. R., & Abdel-Rahman, E. M. (2012). A Metaheuristic Bat-Inspired Algorithm for Full Body Human Pose Estimation, In *Proceedings of Ninth Conference on Computer and Robot Vision* (pp. 369-375). doi:10.1109/CRV.2012.55

Alaguna, C., & Gomez, J. (2018). Maze benchmark for testing evolutionary algorithms. In *Proceedings of the Genetic and Evolutionary Computation Conference Companion* (pp. 1321-1328). doi:10.1145/3205651.3208285

Aljarah, I., Faris, H., & Mirjalili, S. (2018). Optimizing connection weights in neural networks using the whale optimization algorithm. *Soft Computing*, *1*(22).

Aziz, M. A. E., Ewees, A. A., & Hassanien, A. E. (2017). Whale Optimization Algorithm and Moth-Flame Optimization for multilevel thresholding image segmentation. *Expert Systems with Applications*, *I*(83), 242–256. doi:10.1016/j.eswa.2017.04.023

Belkhiri, Y., Kamel, N., & Drias, H. (2022). An efficient multi-swarm elephant herding optimization for solving community detection problem in complex environment. *Concurrency and Computation*, *34*(3). Advance online publication. doi:10.1002/cpe.6590

Bendimerad, L. S., & Drias, H. (2021). An Artificial Orca Algorithm for Continuous Problems. In A. Abraham, T. Hanne, O. Castillo, N. Gandhi, T. Nogueira Rios, & T. P. Hong (Eds.), *Hybrid Intelligent Systems. HIS 2020. Advances in Intelligent Systems and Computing* (Vol. 1375). Springer. doi:10.1007/978-3-030-73050-5_68

Craik, A. D. D. (2004). The origins of water wave theory. *Annual Review of Fluid Mechanics*, *I*(36), 1–28. doi:10.1146/annurev.fluid.36.050802.122118

Dorigo, M., & Stutzle, T. (2006). Ant Colony Optimization. *IEEE Computational Intelligence Magazine*, *1*(4), 28–39. doi:10.1109/MCI.2006.329691

Dutfield, S. (2022). *Killer Whale Hunting Strategies*. Academic Press.

Gupta, S., Singh, V. P., Singh, S. P., Prakash, T., & Rathore, N. S. (2016). Elephant herding optimization based PID controller tuning. *Int. J. Adv. Technol. Eng. Explor*, *I*(3), 194–198. doi:10.19101/IJATEE.2016.324005

Heraguemi, K., Kamel, N. & Drias, H. (2018). Multi-objective bat algorithm for mining numerical association rules. *International Journal of Bio-Inspired Computation, 1*(11).

Houacine, N. A., & Drias, H. (2021). Discrete Elephant Algorithms for Target Detection in Complex and Unknown Environments. In M. R. Senouci, M. E. Y. Boudaren, F. Sebbak, & M. Mataoui (Eds.), *Advances in Computing Systems and Applications. CSA 2020. Lecture Notes in Networks and Systems* (Vol. 199). Springer. doi:10.1007/978-3-030-69418-0_9

Khennak, I., & Drias, H. (2017). Bat-Inspired Algorithm Based Query Expansion for Medical Web Information Retrieval. *Journal of Medical Systems*, *1*(41).

Killer-Whale-Org. (2022). *Killer Whale Habitat and Distribution*. https://www.killer-whale.org

Mafarja, M. M., & Mirjalili, S. (2017). Hybrid Whale Optimization Algorithm with simulated annealing for feature selection. *Neurocomputing*, *I*(260), 302–312. doi:10.1016/j.neucom.2017.04.053

Mandal, S. (2018). Elephant swarm water search algorithm for global optimization. *Sadhana, 1*(1).

Mirjalili, S., & Lewis, A. (2016). The Whale Optimization Algorithm. *Advances in Engineering Software*, *I*(95), 51–67. doi:10.1016/j.advengsoft.2016.01.008

Mishra, S., & Bande, P. (2008). Maze Solving Algorithms for Micro Mouse. *Proceedings of the IEEE International Conference on Signal Image Technology and Internet Based Systems*, 86-93. doi:10.1109/SITIS.2008.104

Mohsin, S. M., Javaid, N., Madani, S. A., Akber, S. M. A. S. M. A., Manzoor, S., & Ahmad, J. (2018). Implementing Elephant Herding Optimization Algorithm with different Operation Time Intervals for Appliance Scheduling in Smart Grid. *Proceedings of the 32nd International Conference on Advanced Information Networking and Applications Workshops*, 240-249.

Oliveira, M., Pinheiro, D., Andrade, B., Bastos-Filho, C., & Menezes, R. (2016). Communication Diversity in Particle Swarm Optimizers. *Proceedings of the International Conference On Swarm Intelligence*, 77-88. doi:10.1007/978-3-319-44427-7_7

OrcaLab. (2022). *Orca Social Organization*. https://orcalab.org/orcas/orca-social-organization

Passino, K. M. (2002). Biomimicry of Bacterial Foraging for Distributed Optimization And Control. IEEE Control Systems Magazine, 52-67.

Sea-World Parks and Entertainment. (2022). *All About Killer Whales-Communication and Echolocation*. Author.

Sood, M., & Bansal, S. (2013). K-Medoids Clustering Technique using Bat Algorithm. *International Journal of Applied Information Systems*, *I*(5), 20–22. doi:10.5120/ijais13-450965

Tuba, E., Alihodzic, A., & Tuba, M. (2017). Multilevel image thresholding using elephant herding optimization algorithm. *Proceedings of the 14th International Conference on Engineering of Modern Electric Systems (EMES)*, 240-243. doi:10.1109/EMES.2017.7980424

Tuba, E., & Stanimirovic, Z. (2017) Elephant herding optimization algorithm for support vector machine parameters tuning. *Proceedings of the 9th International Conference on Electronics, Computers and Artificial Intelligence (ECAI)*, 1-4. doi:10.1109/ECAI.2017.8166464

Wang, G. G., Dos Santos Coelho, L., Gao, X. Z., & Deb, S. (2016). A new metaheuristic optimization algorithm motivated by elephant herding behavior. *International Journal of Bio-inspired Computation*, *I*(8), 394. doi:10.1504/IJBIC.2016.081335

Yang, X. S., & He, X. (2013). Bat Algorithm: Literature Review and Applications. *International Journal of Bio-inspired Computation*, *I*(5), 141–149. doi:10.1504/IJBIC.2013.055093

Yang, X. S., & He, X. (2013). Firefly Algorithm: Recent Advances and Applications. *Int. J. Swarm Intelligence.*, *I*(1), 36–50. doi:10.1504/IJSI.2013.055801

*Habiba Drias received the M.S. degree in computer science from CWRU Cleveland OHIO USA in 1984 and the Ph.D. degree in computer science from USTHB, Algiers, Algeria in collaboration with UPMC Paris, France, in 1993. She is a full professor at USTHB since 1999 and directs the Laboratory of Research in Artificial Intelligence (LRIA). By the past, she was the head of the Computer Science Institute of USTHB and the general director of the Algerian National Institute of Informatics. She has published more than 200 papers in well-recognized international conference proceedings and journals and has directed 25 Ph.D. theses, 38 master theses and 31 engineer projects. In 2013, she won the Algerian Scopus award in computer science and in 2015, she was selected by a jury of international academicians as a funding member of the Algerian Academy of Science and Technology (AAST).*

*Bendimerad Lydia Sonia, in her second year as PhD at the University of Science and Technology of Houari Boumediene in Algeria, is passionate about research, modernization and new technologies.*

*Yassine Drias received his PhD degree in artificial intelligence from the University of Milano-Bicocca in 2017. Prior to that, he prepared his Master's degree in Intelligent Informatics Systems at the University of Sciences and Technologies of Algiers in collaboration with ENSMA Poitiers, France and graduated in 2013. He is currently working on topics including meta-heuristics, Web information foraging, Multiagent Systems, Bio-inspired computing and Data Mining. His works have appeared in computer science journals and international conferences proceedings.*